\pdfoutput=1
\documentclass[sn-mathphys]{sn-jnl}% Math and Physical Sciences Reference Style
\jyear{2021}%

\theoremstyle{thmstyleone}%
\theoremstyle{thmstyletwo}%

\theoremstyle{thmstylethree}%

\usepackage[justification=centering]{caption}
\newcommand{\tabincell}[2]{\begin{tabular}{@{}#1@{}}#2\end{tabular}}
\begin{document}

\title[SAL-CNN]{SAL-CNN: Estimate the Remaining Useful Life of Bearings Using Time-frequency Information}

%\author{Bingguo Liu\inst{1} \and Zhuo Gao\inst{2}}
\author[1]{\fnm{Bingguo} \sur{Liu}}\email{Liu\_bingguo@hit.edu.cn}

\author[1]{\fnm{Zhuo} \sur{Gao}}\email{gaozhuox@foxmail.com}
% %\equalcont{These authors contributed equally to this work.}

 \author*[1]{\fnm{Binghui} \sur{Lu}}\email{miknet0594@163.com}
% %\equalcont{These authors contributed equally to this work.}

\author[1]{\fnm{Hangcheng} \sur{Dong}}\email{hunsen\_d@hit.edu.cn}
% %\equalcont{These authors contributed equally to this work.}

\author[2]{\fnm{Zeru} \sur{An}}\email{sxwtazr@163.com}
% %\equalcont{These authors contributed equally to this work.}

 \affil[1]{\orgdiv{School of Instrumentation Science and Engineering}, \orgname{Harbin Institute of Techonoloy}, \orgaddress{\city{Harbin}, \postcode{150001},  \country{China}}}

 \affil[2]{\orgname{Shanghai Spaceflight Precision Machinery Institute}, \orgaddress{\city{Shanghai}, \postcode{201600},  \country{China}}}
%\institute{School of Instrumentation Science and Engineering, Harbin Institute of Techonoloy, Harbin, 150001, China}


\abstract{In modern industrial production, the prediction ability of remaining useful life (RUL) of bearings directly affects the safety and stability of the system. Traditional methods require rigorous physical modeling and perform poorly for complex systems. In this paper, an end-to-end RUL prediction method is proposed, which uses short-time Fourier transform (STFT) as preprocessing. Considering the time correlation of signal sequences, a long and short-term memory network is designed in CNN, incorporating the convolutional block attention module, and understanding the decision-making process of the network from the interpretability level. Experiments were carried out on the 2012PHM dataset and compared with other methods, and the results proved the effectiveness of the method.}




\maketitle

\section{Introduction}\label{sec1}

In the operation of equipment, a large number of faults are caused by bearing failure. Prediction of remaining useful life (RUL) of bearings has become a key technology to ensure mechanical work safety. Existing bearing RUL prediction methods include two types, statistical life model prediction and data-driven prediction. Model-based methods include particle filtering\cite{bib1}, Eyring model\cite{bib2}, Weibull distribution\cite{bib3}, etc. These methods need a large number of statistical data as the basis to have certain reliability, but it is difficult to establish an accurate and general mathematical model for complex equipment.

The data-driven method uses the end-to-end training strategy and uses the state monitoring data to predict. A.Soualhi et al. proposed a rolling bearing condition monitoring method combining Hilbert-Huang transform, support vector machine, and support vector regression\cite{bib4}. S.A. Aye et al. proposed an optimal Gaussian process regression (GPR) for low-speed bearing RUL\cite{bib5}. Zhang Z et al. converted these signals into the frequency domain using wavelet packet decomposition and fast Fourier transform and trained artificial neural network (ANN)\cite{bib6}.

In the field of CNN, G.S. Babu et al. and L. Ren et al. respectively used fast Fourier transform and wavelet transform as pretreatments and used CNN to predict RUL\cite{bib7}\cite{bib8}. Zhu et al. used the original vibration signal as the input of CNN for training and testing\cite{bib9}. A.Z. Hinchi et al. using CNN and LSTM to predict RUL\cite{bib10}. Biao Wang et al. remember information by feedback connection that feeds back the output of the convolutional layer as input\cite{bib11}. Xiang Li Wei et al. adopt the convolutional neural network of multi-scale feature fusion to gradually extract high-level features and discard low-level features at the same time. Features of different levels are fused, which improves the characteristic of one-way step-by-step transmission of feature information in the traditional convolutional network\cite{bib12}. The difference between this paper and other methods is that the time-frequency diagram of vibration signals is taken as the network input, and CNN integrates LSTM and CBAM.

Focusing on RUL prediction, STFT pretreatment attention and memory based CNN (SAL-CNN) is proposed in this paper. Based on CNN, the short-time Fourier transform (STFT) is used as the vibration signal preprocessing, and long short-term memory(LSTM) is integrated to make up for the deficiency that traditional CNN cannot consider the time correlation between vibration signal sequences, and convolutional block attention module (CBAM) is used to achieve accurate acquisition of fault features and improve the accuracy of network life prediction.

The main contributions of this paper are as follows: A CNN framework (SAL-CNN) is proposed, which takes STFT transformation as pretreatment and integrates CBAM and LSTM; visualizing the output of CBAM to explain the framework, discuss the relationship between RUL and frequency, and enhance experimental credibility.

This paper introduces the basic theory about LSTM and CBAM in Section 2, the preprocessing algorithm and network SAL-CNN are detailed in Section 3, the experimental situation and comparative analysis are introduced in Section 4, and the interpretability of the network is subsequently studied. Finally, the summary is made in Section 5.

\section{Related work}\label{sec2}

\subsection{Long short-term memory}\label{subsec9}
\begin{figure}[h]
\centering
\includegraphics[width=0.8\textwidth]{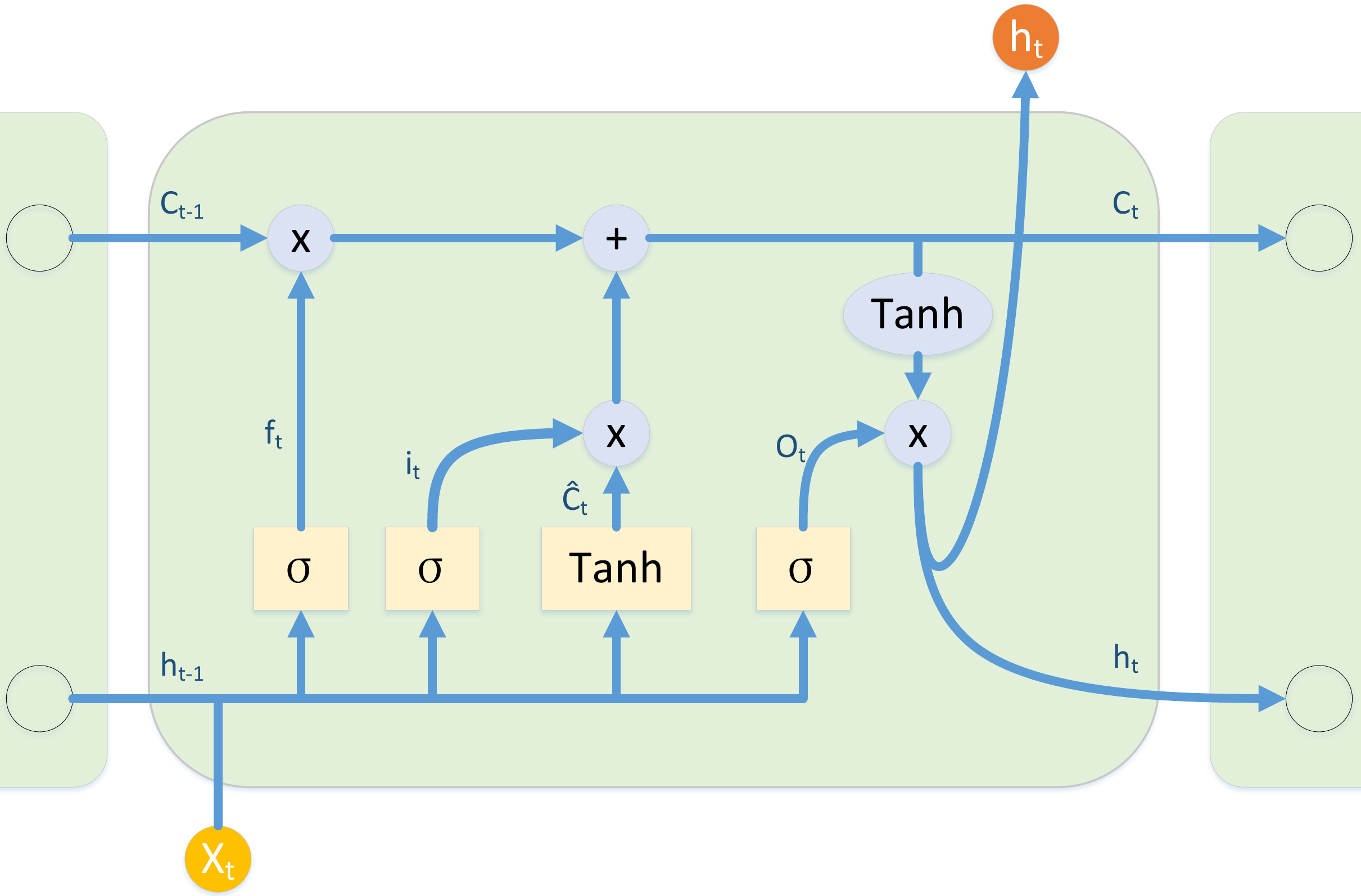}
\caption{LSTM structures}\label{<figure1>}
\end{figure}

The recurrent neural network (RNN)\cite{bib13} takes both the output characteristics of the last moment and the data of the current moment as input.  But gradient disappearance and gradient explosion will occur with the increase of cycle layers. LSTM\cite{bib14} is a special RNN, it introduces the concepts of gating unit and cell state. The input gate controls the output information from the previous layer, and the memory information from the previous moment is controlled by the forgetting gate. The most common LSTM structures are shown in Fig.\ref{<figure1>}.

The expression of the forgetting gate is:
\begin{equation}
f_t=\sigma\left(W_f\cdot\left[h_{t-1},x_t\right]+b_f\right)\label{eq1}
\end{equation}

The expression of the input gate is:
\begin{equation}
i_t=\sigma\left(W_i\cdot\left[h_{t-1},x_t\right]+b_i\right)\label{eq2}
\end{equation}
\begin{equation}\widetilde{C_t}=tanh{\left(W_C\cdot\left[h_{t-1},x_t\right]+b_C\right)}\label{eq3}
\end{equation}

The expression of the output gate is:
\begin{equation}
o_t=\sigma\left(W_o\cdot\left[h_{t-1},x_t\right]+b_o\right)\label{eq4}
\end{equation}
\begin{equation}
h_t=o_t\cdot tanh{\left(C_t\right)}\label{eq5}
\end{equation}

Where $W_f,W_i,Wc$ and $W_o$ represent the weight matrix respectively, $\sigma$ represents the sigmoid activation function, and  act on   after passing through the forgetting gate and the input gate. The expression is:
\begin{equation}
C_t=f_t\cdot C_{t-1}+i_t\cdot{\widetilde{C}}_{t-1}\label{eq6}
\end{equation}

\subsection{Convolutional block attention module}\label{subsec2}

\begin{figure}[h]
\centering
\includegraphics[width=1.0\textwidth]{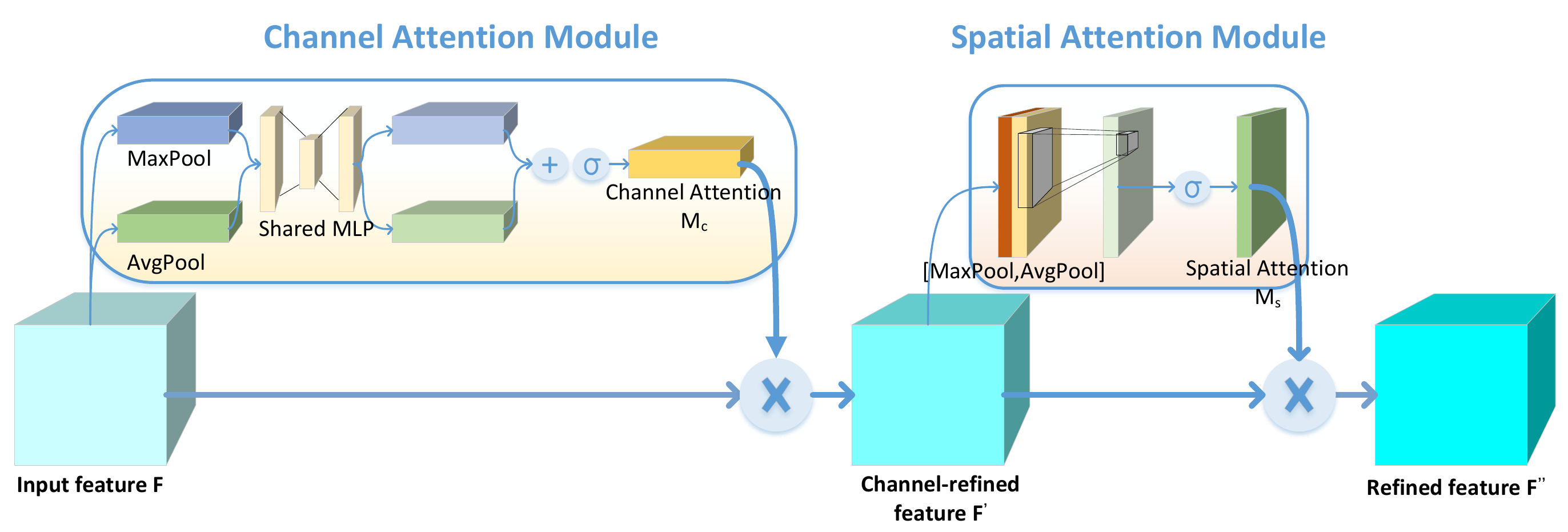}
\caption{CBAM structures}\label{<figure2>}
\end{figure}

Convolutional block attention module (CBAM)\cite{bib15} is a simple and effective injection module for feedforward convolutional neural network, its structure is shown in the Fig.\ref{<figure2>}. As a lightweight module, CBAM can be placed behind any feature graph as required. For feature map $F\in R_c\cdot h\cdot w$ of an intermediate layer, CBAM will deduce the 1-dimensional channel attention map $M_c\in R_c\cdot 1\cdot 1$ and 2-dimensional spatial attention map $M_s\in R1\cdot H\cdot W$, as shown below :
\begin{equation}
F’=M_c(F)\otimes F\label{eq7}
\end{equation}
\begin{equation}
F''=M_s(F')\otimes F'\label{eq8}
\end{equation}

Where $\otimes$ is element-wise multiplication, and the expressions of the channel attention module Mc(F) and space attention module Ms(F) are as follows :
\begin{equation}
M_C\left(F\right)=\sigma\left(\mathrm{MLP}\left(\mathrm{AvgPool}\left(F\right)\right)+\mathrm{MLP} \left(\mathrm{MaxPool}\left(F\right)\right)\right)\label{eq9}
\end{equation}
\begin{equation}
M_S\left(F\right)=\sigma\left(f\left(\left[\mathrm{AvgPool}\left(F\right),\mathrm{MaxPool} \left(F\right)\right]\right)\right)\label{eq10}
\end{equation}

Where $\sigma$ is Sigmoid activation function, and $f$ is convolution operation.

\section{Framework}\label{sec3}

\begin{figure}[h]
\centering
\includegraphics[width=0.9\textwidth]{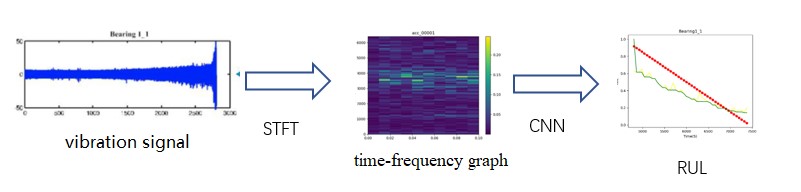}
\caption{Flow diagram}\label{<figure3>}
\end{figure}

Aiming at the problems that existing models cannot accurately predict RUL and work in real-time, an end-to-end detection framework, SAL-CNN, is proposed. The flow diagram is shown in Fig.\ref{<figure3>}. The method includes two parts: vibration signal preprocessing and convolution neural network. CNN is designed to integrate convolutional block attention module and long and short term memory network.

\subsection{Vibration signal preprocessing}\label{subsec3}

Since vibration signals have many spectral components and are non-stationary signals, short-time Fourier transform\cite{bib16} is used to preprocessing the input data to obtain the time-frequency graph diagram. Short-time Fourier transform is a kind of joint time-frequency analysis method. Based on Fourier transform, the whole segment of signal is processed by window segmentation, and each small segment is considered as a stationary signal. The window function slips along the time axis, and the Fourier transform is carried out in the neighborhood at any time.

For non-stationary signals $x\left(t\right)\in L^2\left(R\right)$, the short-time Fourier transform of $x\left(t\right)$ is expressed as:
\begin{equation}
STFT_x\left(t,f\right)=\int_{-\infty}^{\infty}x\left(\tau\right)h\left(\tau-t\right)e^{-j2\pi ft}d\tau\label{eq11}
\end{equation}

Where $h\left(\tau-t\right)$ is window function.

Take the Fourier transform of $h\left(t\right)$, the energy is concentrated at the low frequency range, so it's usually thought of a low-pass filter. As can be seen from the equation(11), window function $h\left(t\right)$ moves along the time axis and conducts segmented processing on $h\left(t\right)$ in STFT transformation, the expression is:
\begin{equation}
x_t\left(\tau\right)=x\left(\tau\right)h\left(\tau-t\right)\label{eq12}
\end{equation}

Later, take the Fourier transform of $x_t\left(\tau\right)$.

After windowing the signal, all the features of the signal covered by the window function will be displayed. The selection of the window function directly affects the results after STFT transformation. The frequency resolution is mainly determined by the main lobe width, so the window function with the narrowest main lobe and the smallest sidelobe peak value should be selected. The common window functions are rectangular window, triangular window, Hamming window, Hamming window, Blackman window, etc., and hamming window is selected as the window function comprehensively.

\begin{figure}[h]
\centering
\includegraphics[width=0.8\textwidth]{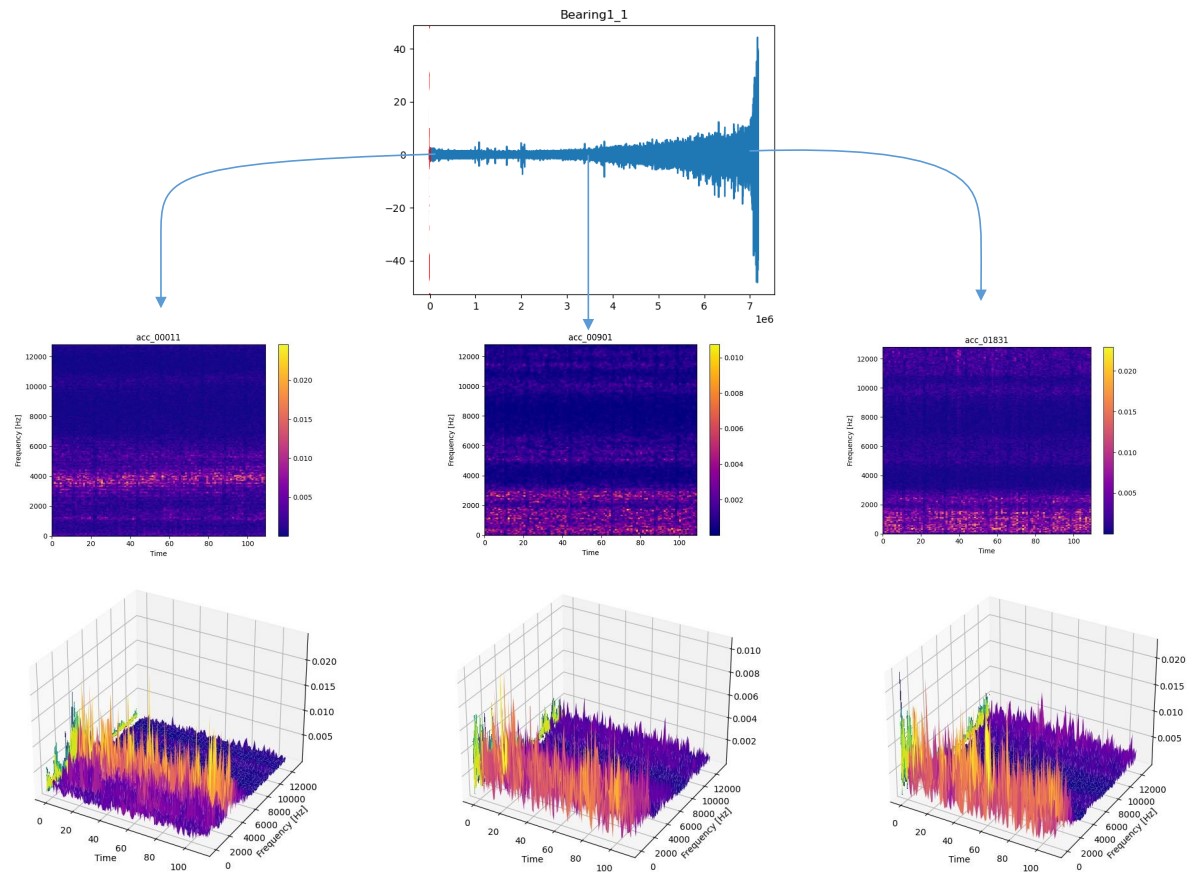}
\caption{Three stages of vibration signal}\label{<figure4>}
\end{figure}

Fig.\ref{<figure4>} shows the vibration signal of bearing 1-1 in the whole life cycle and the time-frequency diagram after STFT transformation in the initial, middle and final stages, the first row is the vibration signal diagram, and the second row is the time-frequency diagram after STFT transformation. The X-axis represents the time axis, and the Y-axis represents the vibration frequency. The more brightly colored part represents the more concentrated energy of the vibration frequency in the same time series. The third row is a 3D display of the time-frequency graph, where the z-axis represents the frequency occurrence times. As can be seen from the figure, with the degradation of bearings and the increase of vibration signal amplitude, signals gradually concentrated in the low-frequency range. By this time, the high-frequency part was mainly an external noise signal, which made little contribution to life prediction. Therefore, the frequency of the upper part of the vibration signal was filtered out, and the remaining time-frequency information was used to predict RUL.

\subsection{Network model structure}\label{subsec4}

\begin{figure}[h]
\centering
\includegraphics[width=0.8\textwidth]{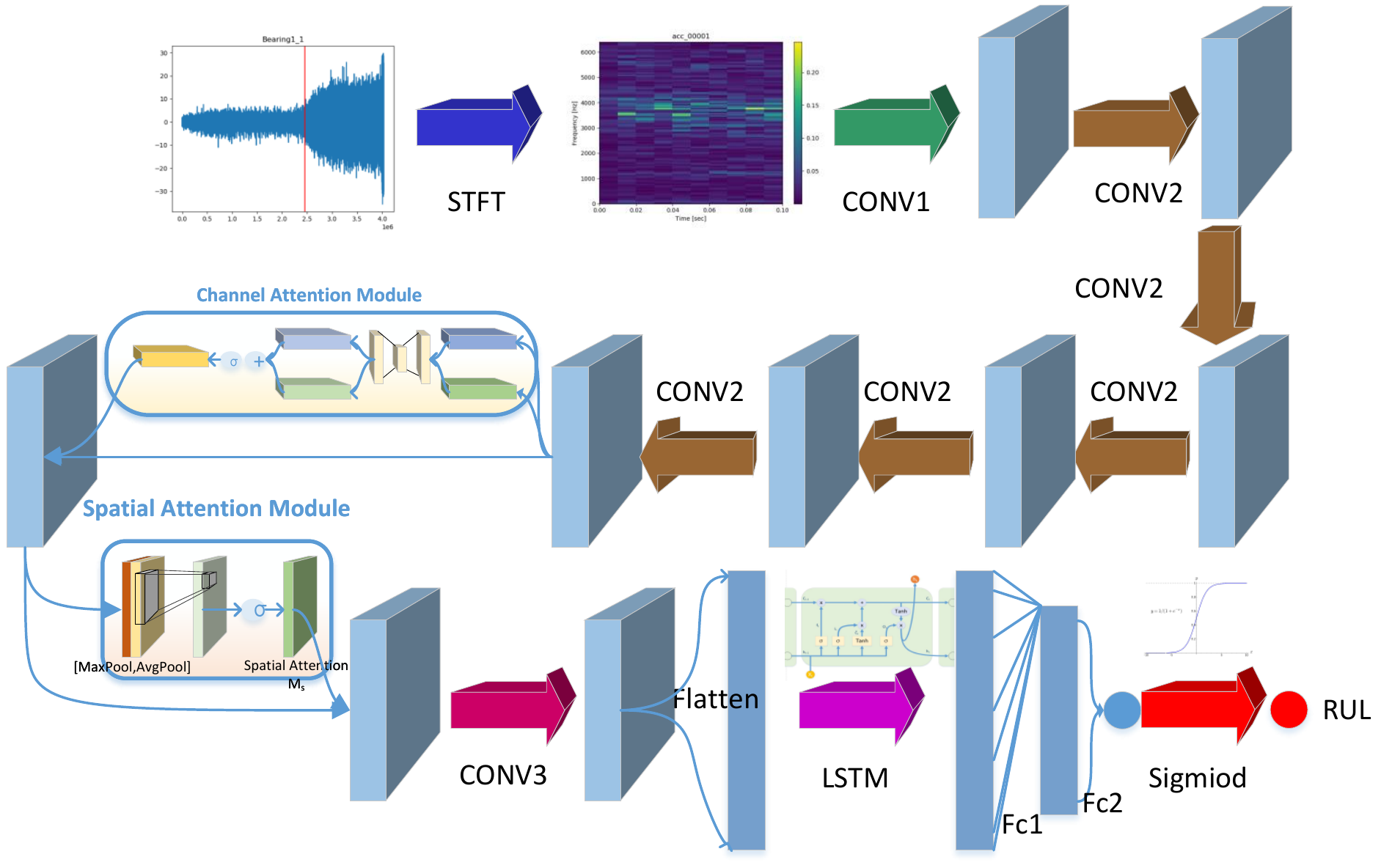}
\caption{Structure of SAL-CNN}\label{<figure5>}
\end{figure}

Considering that the input data is time-dependent, it is proposed to integrate a convolutional neural network and long and short term memory network. On this basis, a convolutional block attention module is introduced to improve the learning performance of the network, then visualizing the output of this module. The structure of the proposed neural network is shown in Fig.\ref{<figure5>}.

Firstly, the vibration signal is preprocessed by STFT transformation, the size of time-frequency diagram is 11$\times$129, after preprocessing, enters into a six-layer convolution operation, and then into CBAM. After that, a one-layer convolution operation with a convolution kernel is used to compress the information, followed by LSTM, the last layer is full connection layer and outputs a predicted life value, Table \ref{tab:table1} describes specific network parameters.

\begin{table}[h]
\centering
\caption{Network parameters of SAL-CNN}
\label{tab:table1}
\resizebox{0.6\textwidth}{45mm}{
\begin{tabular}{cll}
\toprule
\multicolumn{1}{l}{Layer} & Parameter    & Value    \\ \midrule
\multirow{3}{*}{STFT}     & window       & Hamming  \\
                          & fs           & 25600 Hz \\
                          & Nperseg      & 512      \\
\multirow{3}{*}{Conv1}    & In channels  & 1        \\
                          & Kernel size  & 3×3      \\
                          & Out channels & 5        \\
\multirow{3}{*}{Conv2}    & In channels  & 5        \\
                          & Kernel size  & 3×3      \\
                          & Out channels & 5        \\
\multirow{3}{*}{Conv3}    & In channels  & 5        \\
                          & Kernel size  & 3×3      \\
                          & Out channels & 1        \\
\multirow{3}{*}{LSTM}     & Input size   & 1408     \\
                          & Hidden size  & 1408     \\
                          & Num layers   & 2        \\ \botrule
\end{tabular}}
\end{table}

\section{Experiments and discussions}\label{sec4}

\subsection{Data sets and evaluation indicators}\label{subsec5}

In order to verify the effectiveness of the method, the data set of the 2012PHM Challenge\cite{bib17} was used as the validation data set. The data set was provided by the FEMTO-ST research institute in France and was obtained by building an experimental platform in a laboratory environment. Three working conditions were used for experiments: (1)rotate speed:1800 r/min, radial force: 4000 N; (2)rotate speed: 1650 r/min, radial force: 4200 N; (3)rotate speed: 1500 r/min, radial force: 5000 N. The experimental platform collects vibration signals of the whole life cycle of bearings, and a total of 17 bearings are tested and collected vibration signals,Table \ref{tab:table2} is the distribution of bearings.

\begin{table}[h]
\centering
\caption{distribution of bearings in 2012PHM Challenge}
\label{tab:table2}
\begin{tabular}{@{}ccccc@{}}\toprule
operating mode     & {radial force(N)}        & rotate speed(rpm)        & Bearing                                                                      \\\midrule
\multirow{2}{*}{1} & \multirow{2}{*}{4000}     &{\multirow{2}{*}{1800}}& Bearing1-1,Bearing1-2\\&&&    Bearing1-3,Bearing1-4\\
                   &                           &                       & Bearing1-5,Bearing1-6,Bearing1-7\\
\multirow{2}{*}{2} & \multirow{2}{*}{4200}     & \multirow{2}{*}{1650} & Bearing2-1,Bearing2-2\\&&&
Bearing2-3,Bearing2-4\\
			   &                           &                       & Bearing2-5,Bearing2-6,Bearing2-7 \\
\multirow{2}{*}{3} & \multirow{2}{*}{5000}     & \multirow{2}{*}{1500} & \multirow{2}{*}{Bearing3-1,Bearing3-2,Bearing3-3}\\
                   &                           &                       &                                  \\\botrule                                          
\end{tabular}
\end{table}

In order to quantitatively evaluate the network structure, MAE is used as the evaluation index. As shown by expression 13, the more accurate the prediction results are, the lower MAE is:
\begin{equation}
MAE=\frac{1}{s}\sum_{i=1}^{s}\left({\rm ActRUL}_i-{\rm PreRUL}_i\right)\label{eq13}
\end{equation}
where S is the number of testing samples, ActRULi is the actual RUL values corresponding to the i-th testing sample, and PreRULi is the predictive RUL values corresponding to it.

\subsection{Experimental setup}\label{subsec6}

All bearings were used as prediction bearings in turn, and the remaining 16 bearings were used as training sets in each experiment to obtain cross-validation results.

The training set is input into SAL-CNN for training, SAL-CNN is implemented by PyTorch with Nvidia GeForce GTX 2080ti GPU, the iteration training is terminated after 150 epochs. The loss function is L1 loss and the optimizer is Adam, the learning rate is set to 0.001 and batch size is 32, dropout rate is 0.1, The number of LSTM cycle layers is 1.

\subsection{Analysis}\label{subsec7}

\begin{figure}[h]
\centering
\includegraphics[width=0.5\textwidth]{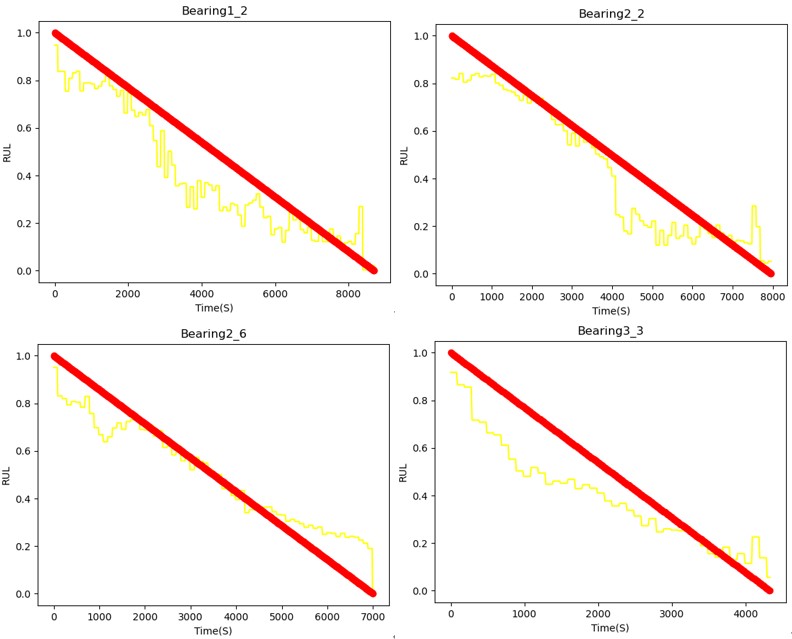}
\caption{RUL prediction results}\label{<figure6>}
\end{figure}

Fig.\ref{<figure6>} shows the prediction results of some bearings. The red curve is the real RUL and the yellow curve is the predicted RUL. It can be seen that the yellow curve basically fits the true value.

This paper conducted comparative experiments with 4 methods on the 2012PHM Challenge data set. Table \ref{tab:table3} shows MAE indexes compared with other methods. DNN is also known as multi-layer perceptron (MLP), Single Scale-Low (SSL), and Single Scale-High (SSH) use low-scale features and high-level features respectively for RUL estimation without feature concatenation, and multi-scale method integrates all scale  feature\cite{bib18}.

The comparison result indicates that the MAE obtained by the proposed algorithm is higher than others. Among the 17 bearings, the prediction results of 11 bearings are best than others, and 2/3 of the results are better and evenly distributed under each working condition, indicating that this method is effective for bearings under three working conditions. According to the average value of the final MAE, it can be seen that the results in this paper are improved by 2.8 compared with the multi-scale method.

\begin{table}[h]
\centering
\caption{Comparison of life prediction results based on MAE index}
\label{tab:table3}
\begin{tabular}{@{}cccccc@{}}\toprule
Bearing    & DNN  & SSL  & SSH  & multi-scale\cite{bib18} & SAL-CNN \\\midrule
Bearing1-1 & 30.4 & 24.3 & 24.4 & 21.4        & \textbf{16.9}    \\
Bearing1-2 & 28.5 & 17.4 & 22.4 & 14.7        & \textbf{13.0}    \\
Bearing1-3 & 16.3 & 9.5  & 9.3  & \textbf{7.8}         & 10.3    \\
Bearing1-4 & 32.1 & \textbf{21.3} & 24.5 & 21.8        & 23.0    \\
Bearing1-5 & 28.7 & 21.4 & 19.5 & 18.5        & \textbf{11.8}    \\
Bearing1-6 & 32.4 & 24.5 & 20.4 & \textbf{20.3}        & 21.4    \\
Bearing1-7 & 16.6 & 9.6  & 11.7 & \textbf{8.3}         & 11.9    \\
Bearing2-1 & 28.6 & 24.2 & 25.2 & 21.5        & \textbf{18.5}    \\
Bearing2-2 & 26.4 & 15.3 & 16.5 & 16.2        & \textbf{8.8}     \\
Bearing2-3 & 40.5 & 35.2 & 31.5 & 32.8        & \textbf{14.8}   \\
Bearing2-4 & 22.3 & 11.6 & 9.9  & \textbf{6.1}         & 24.2    \\
Bearing2-5 & 31.4 & 27.3 & 25.2 & 23.6        & \textbf{23.3}    \\
Bearing2-6 & 29,4 & 23.2 & 19.8 & 20.7        & \textbf{8.6}     \\
Bearing2-7 & 38.0 & 28.2 & 28.4 & 27.4        & \textbf{27.1}    \\
Bearing3-1 & 28.3 & 28.5 & 26.3 & \textbf{19.6}        & 25.2    \\
Bcaring3-2 & 30.4 & 25.2 & 23.7 & 23.1        & \textbf{18.1}    \\
Bearing3-3 & 38.5 & 35.4 & 33.5 & 30.4        & \textbf{11.1}    \\\botrule
\end{tabular}
\end{table}

\subsection{Feature information visualization}\label{subsec8}

Feature visualization is a post-hoc interpretation of the pre-training model. After training, the decision-making process of CNN is explained in the form of images, which can intuitively reflect the key areas of input features that the network focuses on.

Table \ref{tab:table4} shows the time-frequency graph and heatmap of bearings 1-1, 1-3,1-7 in the early, middle, and late stages. In time-frequency graph, the ordinate is the frequency and the value range is [0,6400], and the abscissa is time and the value range is [0,1].  In the trained network, visualizing the output of the CBAM layer, and obtaining the heatmap, where the highlighted parts represent the areas that the network is more concerned with.  As time goes by, the frequency concentrated range in the time-frequency graph is from 3000-4000Hz to 0-2000Hz, and the high-frequency part is basically no energy. It can also be seen from the heatmap that the network pays more attention to the information of the low-frequency part in the late stage, it means low-frequency signal may become more import. 

\begin{table}[h]
  
  \caption{Spectrogram and Heatmap of bearing in three stages}
  \label{tab:table4}
  \centering
  \begin{tabular}{  c  c  c  c  c  }
    \toprule
    bearing & Image type & Early stage &Middle stage & Late stage \\\midrule
	\multirow{8}{*}{1-1}&\tabincell{c}{time-frequency\\graph}&
    	\begin{minipage}[b]{0.2\columnwidth}
		\centering
		\raisebox{-.5\height}{\includegraphics[width=80pt,height=80pt]{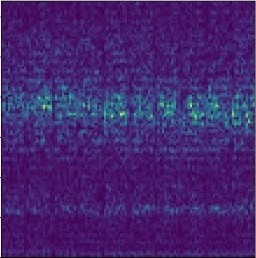}}
	\end{minipage}
    & \begin{minipage}[b]{0.2\columnwidth}
		\centering
		\raisebox{-.5\height}{\includegraphics[width=80pt,height=80pt]{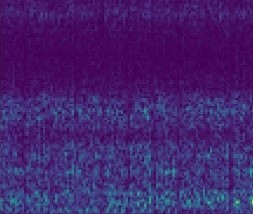}}
	\end{minipage}
    & \begin{minipage}[b]{0.2\columnwidth}
		\centering
		\raisebox{-.5\height}{\includegraphics[width=80pt,height=80pt]{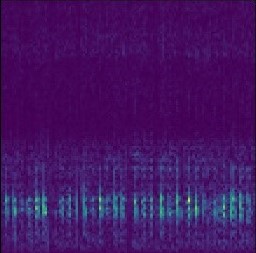}}
	\end{minipage}
    \\
  
        & Heatmap &
    \begin{minipage}[b]{0.2\columnwidth}
		\centering
		\raisebox{-.5\height}{\includegraphics[width=80pt,height=80pt]{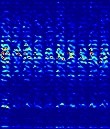}}
	\end{minipage}

    & \begin{minipage}[b]{0.2\columnwidth}
		\centering
		\raisebox{-.5\height}{\includegraphics[width=80pt,height=80pt]{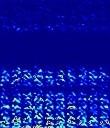}}
	\end{minipage}
    & \begin{minipage}[b]{0.2\columnwidth}
		\centering
		\raisebox{-.5\height}{\includegraphics[width=80pt,height=80pt]{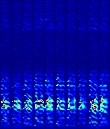}}
	\end{minipage}\\\

\multirow{8}{*}{1-3}&\tabincell{c}{time-frequency\\graph}&
    	\begin{minipage}[b]{0.2\columnwidth}
		\centering
		\raisebox{-.5\height}{\includegraphics[width=80pt,height=80pt]{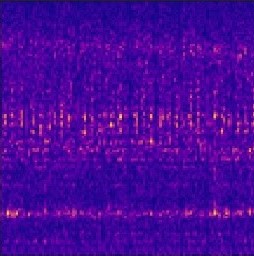}}
	\end{minipage}
    & \begin{minipage}[b]{0.2\columnwidth}
		\centering
		\raisebox{-.5\height}{\includegraphics[width=80pt,height=80pt]{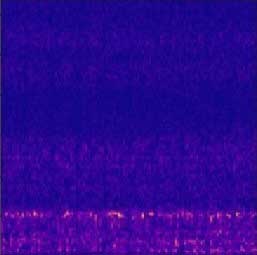}}
	\end{minipage}
    & \begin{minipage}[b]{0.2\columnwidth}
		\centering
		\raisebox{-.5\height}{\includegraphics[width=80pt,height=80pt]{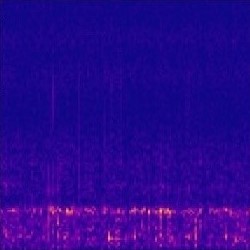}}
	\end{minipage}
    \\
  
        & Heatmap &
    \begin{minipage}[b]{0.2\columnwidth}
		\centering
		\raisebox{-.5\height}{\includegraphics[width=80pt,height=80pt]{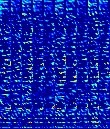}}
	\end{minipage}

    & \begin{minipage}[b]{0.2\columnwidth}
		\centering
		\raisebox{-.5\height}{\includegraphics[width=80pt,height=80pt]{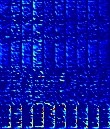}}
	\end{minipage}
    & \begin{minipage}[b]{0.2\columnwidth}
		\centering
		\raisebox{-.5\height}{\includegraphics[width=80pt,height=80pt]{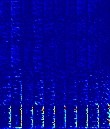}}
	\end{minipage}\\

\multirow{8}{*}{1-5}&\tabincell{c}{time-frequency\\graph}&
    	\begin{minipage}[b]{0.2\columnwidth}
		\centering
		\raisebox{-.5\height}{\includegraphics[width=80pt,height=80pt]{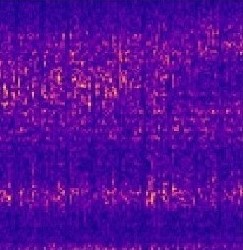}}
	\end{minipage}
    & \begin{minipage}[b]{0.2\columnwidth}
		\centering
		\raisebox{-.5\height}{\includegraphics[width=80pt,height=80pt]{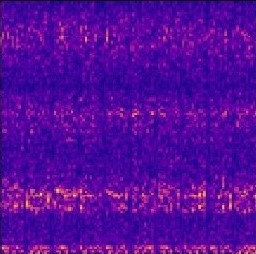}}
	\end{minipage}
    & \begin{minipage}[b]{0.2\columnwidth}
		\centering
		\raisebox{-.5\height}{\includegraphics[width=80pt,height=80pt]{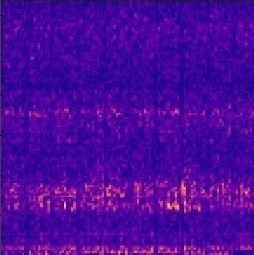}}
	\end{minipage}
    \\
  
        & Heatmap &
    \begin{minipage}[b]{0.2\columnwidth}
		\centering
		\raisebox{-.5\height}{\includegraphics[width=80pt,height=80pt]{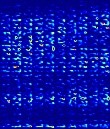}}
	\end{minipage}

    & \begin{minipage}[b]{0.2\columnwidth}
		\centering
		\raisebox{-.5\height}{\includegraphics[width=80pt,height=80pt]{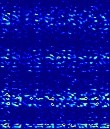}}
	\end{minipage}
    & \begin{minipage}[b]{0.2\columnwidth}
		\centering
		\raisebox{-.5\height}{\includegraphics[width=80pt,height=80pt]{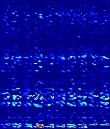}}
	\end{minipage}
    \\ \botrule
  \end{tabular}
  
\end{table}

\section{Summary}\label{sec5}

This paper proposes a long and short-term memory convolutional neural network that introduces a convolutional block attention module. The improved algorithm makes up for the lack of traditional convolutional networks that cannot consider the time correlation of vibration signal sequences. At the same time, the convolutional block attention module further improves the prediction performance of the network. The test on the PHM2012 data set proves that the average MAE is slightly higher than the best recent results.

In addition, we found that for predicting the RUL task, low-frequency vibration signals could be more worthy of attention at a later stage, from the time-frequency graph can directly see the most energy distribute in low-frequency range at a later stage . Using heatmap visualization the output of CBAM, also found that the network at the later stage pay more attention to the low-frequency range, from the aspect of interpretability confirms this conjecture.

\bibliography{sn-bibliography}

\end{document}